\pdfoutput=1

\documentclass[11pt]{article}

\usepackage[final]{acl}

\usepackage{times}
\usepackage{latexsym}
\usepackage{multirow}
\usepackage{booktabs}
\usepackage{enumitem}
\usepackage{placeins}
\usepackage{afterpage}
\usepackage[T1]{fontenc}

\usepackage[utf8]{inputenc}

\usepackage{microtype}

\usepackage{inconsolata}
\usepackage{colortbl} 
\usepackage{xcolor}   

\usepackage{graphicx}
\usepackage{amsmath}
\usepackage[ruled,vlined]{algorithm2e} 
\usepackage{listings}
\usepackage{longtable} 
\usepackage{amssymb}
\usepackage{float}

\lstdefinestyle{sparql}{
  language=SQL,
  keywordstyle=\color{blue}\bfseries,
  commentstyle=\color{gray},
  basicstyle=\ttfamily\small,   
  showstringspaces=false,
  morekeywords={PREFIX,ns,SELECT,WHERE},
  breaklines=true,              
  breakatwhitespace=false,      
  columns=fullflexible          
}

\title{iQUEST: An Iterative Question-Guided Framework for Knowledge Base Question Answering}

\author{
  Shuai Wang \quad \quad \quad \quad \quad Yinan Yu \\
  Department of Computer Science and Engineering \\
  Chalmers University of Technology and University of Gothenburg \\
  SE-41296 Gothenburg, Sweden \\
  \texttt{\{shuaiwa, yinan\}@chalmers.se}
}

\begin{document}
\maketitle
\begin{abstract}
Large Language Models (LLMs) excel in many natural language processing tasks but often exhibit factual inconsistencies in knowledge-intensive settings. Integrating external knowledge resources, particularly knowledge graphs (KGs), provides a transparent and updatable foundation for more reliable reasoning. Knowledge Base Question Answering (KBQA), which queries and reasons over KGs, is central to this effort, especially for complex, multi-hop queries. However, multi-hop reasoning poses two key challenges: (1)~maintaining coherent reasoning paths, and (2)~avoiding prematurely discarding critical multi-hop connections. To tackle these challenges, we introduce iQUEST, a question-guided KBQA framework that iteratively decomposes complex queries into simpler sub-questions, ensuring a structured and focused reasoning trajectory. Additionally, we integrate a Graph Neural Network (GNN) to look ahead and incorporate 2-hop neighbor information at each reasoning step. This dual approach strengthens the reasoning process, enabling the model to explore viable paths more effectively. Detailed experiments demonstrate the consistent improvement delivered by iQUEST across four benchmark datasets and four LLMs. 
The code is publicly available at:
\url{https://github.com/Wangshuaiia/iQUEST}.
\end{abstract}

\section{Introduction}
Large language models (LLMs) have achieved remarkable success across diverse Natural Language Processing (NLP) tasks, yet they often exhibit hallucinations and factual errors in specialized, knowledge-intensive domains~\cite{huang2024survey,martino2023knowledge,minaee2024large}. Fine-tuning LLMs on curated datasets can embed domain-specific knowledge into model parameters, but it is computationally expensive, difficult to update, and opaque in terms of interpretability~\cite{hu2023llm}. As an alternative, Retrieval-Augmented Generation (RAG) techniques query external resources during inference, thereby reducing reliance on repeated retraining~\cite{gao2023retrieval}. Among these resources, knowledge graphs (KGs) offer structured and trustworthy information that is directly verifiable and maintainable, an essential feature for high-stakes scenarios such as healthcare and autonomous driving~\cite{wen-etal-2024-mindmap}.

\begin{figure}[t]
    \centering
    \includegraphics[width=1.0\linewidth]{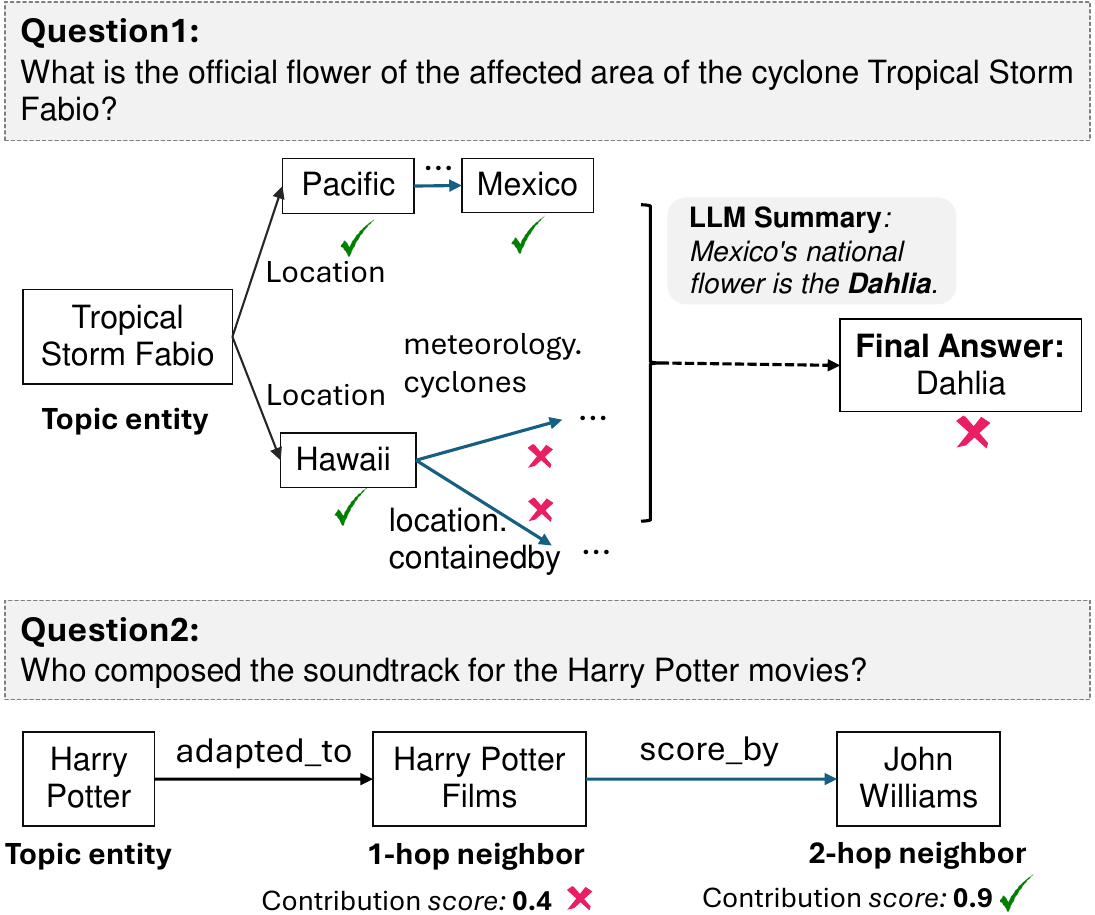}
    \caption{Examples of multi-step reasoning in a knowledge graph, each starting from a topic entity. \textit{Question 1} illustrates how maintaining coherent reasoning paths can be challenging, whereas \textit{Question 2} shows how critical multi-hop connections may be prematurely discarded. Entities marked with \textcolor{green}{\checkmark} are selected in the reasoning path, and those marked with \textcolor{red}{\texttimes} are unselected.}
    \label{fig:intro-example}
    \vspace{-0.5cm}
\end{figure}

In this context, Knowledge Base Question Answering (KBQA) has emerged as a crucial research direction, aiming to query and reason over KGs to answer natural language questions. Complex KBQA tasks often require multi-hop reasoning or multiple constraints~\cite{lan2022complex}, as illustrated by questions like “In the North Pacific region, what is the official flower of the affected area of the cyclone Tropical Storm Fabio?” Answering such queries necessitates iterative exploration of entities, relations, and constraints within large, heterogeneous KGs.

A straightforward way to handle complex queries is the Chain-of-Thought (CoT)~\cite{wei2022chain} framework, which makes each inference step explicit. Building on this idea, some methods treat LLMs as agents that iteratively explore the KG from a topic node (e.g., ToG~\cite{sun2024thinkongraph}, Interactive-KBQA~\cite{xiong-etal-2024-interactive}), while others combine a smaller model with an LLM to gather evidence before synthesizing the final answer (e.g., KG-CoT~\cite{zhao2024kg}). Another direction focuses on sub-question decomposition, either by inserting delimiters into the original query~\cite{huang2023question} or by using an LLM to break down the query for a fine-tuned smaller model~\cite{yixing2024chain}. Although these approaches have shown promise, multi-step reasoning over KGs still face two key challenges.

\textbf{Difficulty in discovering effective reasoning paths.}
Multi-hop inference demands sustained logical focus. However, as reasoning involves multiple interwoven subproblems, models often struggle to maintain direction, losing track after just a few steps. Additionally, ambiguous or tangential entities (e.g., “Mexico” in Question 1 of Figure~\ref{fig:intro-example}) can introduce noise, further disrupting the reasoning path and hindering the identification of the correct answer.

\textbf{Prematurely discarding critical multi-hop connections.}
Existing methods frequently rely on local (1-hop) relations between the question and a topic entity’s neighbors, risking the early elimination of valuable multi-hop paths. For instance, in Question 2 of Figure~\ref{fig:intro-example}, “Harry Potter Films” might be incorrectly discarded due to a lower immediate relevance score, even though its 2-hop neighbor “John Williams” strongly matches the query. Such oversights can derail the entire reasoning process and lead to incorrect answers.

In complex problem-solving scenarios, people often lose focus if their attention is not guided continuously. However, research indicates that by posing and solving smaller questions, humans can maintain higher levels of attention during the reasoning process, thereby improving decision-making and problem-solving~\cite{salmon2021intentional,tofade2013best}. Inspired by this insight, we propose to guide LLMs in multi-step KBQA by iteratively posing simpler sub-questions, thereby maintaining a clear reasoning trajectory over multiple hops.

To discover more robust reasoning paths on KGs, we devise an iterative question-guided framework \textbf{iQUEST} which, at each iteration, generates a sub-question fully answerable from the current context. The LLM then targets that sub-question, retrieves evidence entities, and returns an intermediate answer. Concurrently, we mitigate the brittleness of multi-hop exploration by integrating a Graph Neural Network (GNN) to incorporate semantic information from second-hop neighbors, thereby helping the model look one step ahead and avoid discarding potentially crucial connections.

The key contributions of this paper are summarized as follows:
\begin{itemize}
\item We propose a question-guided reasoning framework that differs from prior work focused solely on question decomposition. Instead of decomposing the original question once, our method enables the LLM to iteratively generate a new sub-question at each reasoning step based on the current state, effectively guiding the reasoning process over the knowledge graph.

\item We design a GNN-based method to aggregate semantic information from second-hop neighbors, allowing the model to look ahead in the KG at each step and enhancing the robustness and accuracy of multi-hop reasoning.

\item We conduct extensive experiments on four benchmark datasets, demonstrating the effectiveness and generalizability of our proposed methods.
\end{itemize}

\begin{figure*}[t]
    \centering
    \includegraphics[width=1.0\linewidth]{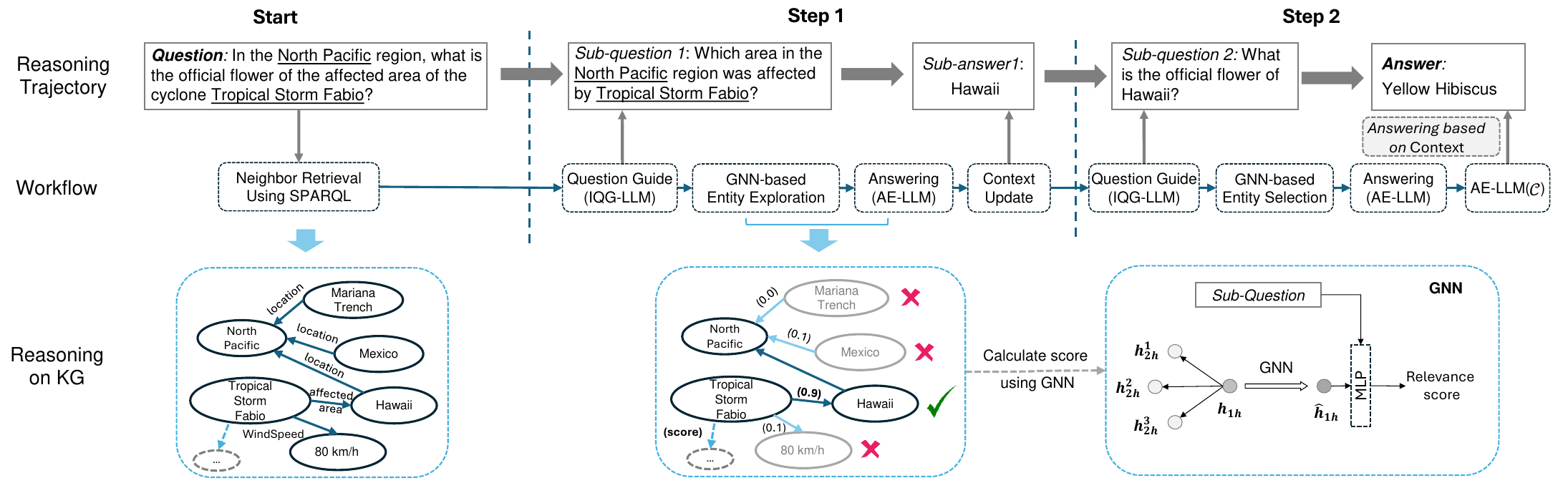}
    \caption{Overview of our framework, illustrating the reasoning trajectory, workflow, and multi-step reasoning process over KGs.}
    \label{fig:main}
\end{figure*}

\section{Related Work} 
\subsection{KBQA} 
Knowledge Base Question Answering (KBQA) aims to leverage a KG to generate answers. Traditional methods are generally IR-based or SP-based: the former retrieves candidate entities and relations from a KG and then ranks them~\cite{saxena2020improving, dai2023fedgamma, chen2022temporal}, whereas the latter converts natural language questions into executable structured queries (e.g., SPARQL)~\cite{das2021case, ye2022rng}.

Recently, LLMs have introduced new possibilities. Several approaches integrate LLMs in different ways: Interactive-KBQA~\cite{xiong-etal-2024-interactive} iteratively refines SPARQL queries; ToG~\cite{sun2024thinkongraph} uses an LLM agent to explore a KG for evidence; FlexKBQA~\cite{li2024flexkbqa} synthesizes training data for smaller models; KG-CoT~\cite{zhao2024kg} fuses small-model reasoning with LLM-based answer finalization. Other work uses tool-like interfaces for external KG reasoning~\cite{jiang2023structgpt} or multi-agent strategies~\cite{zong-etal-2024-triad} to enhance KBQA performance.

\subsection{Question Decomposition} 
A central challenge in complex KBQA is decomposing the query into tractable sub-questions. Early methods used neural architectures (e.g., Seq2Seq or BERT+LSTM) to split questions, but often lacked robustness or altered entity semantics~\cite{zhang2019complex, gu-su-2022-arcaneqa}. Later work introduced explicit delimiters to preserve meaning and sequentially answer sub-questions~\cite{huang2023question}. Increasingly, LLMs have been adopted for decomposition and subsequent reasoning~\cite{yixing2024chain}. However, most focus on segmentation rather than adaptive inference. In contrast, our approach uses targeted intermediate prompts to guide and validate each reasoning step, harnessing the LLM’s capabilities more comprehensively.

\subsection{Entity Exploration} 
Many KBQA systems rely on 1-hop neighbors for candidate retrieval, employing text similarity (e.g., BM25), direct semantic relevance, or MLP-based scoring~\cite{sun2024thinkongraph, xiong-etal-2024-interactive, zhao2024kg, yixing2024chain}. These methods often overlook 2-hop neighbors, potentially missing critical inference paths. Our work addresses this gap by explicitly incorporating 2-hop neighbors into the search process, enabling more thorough exploration and improved answer derivation.

\section{Problem defination}
We consider a KG $\mathcal{G} = \{\langle e, r, e' \rangle \mid e, e' \in \mathcal{E}, r \in \mathcal{R}\}$, where $\mathcal{E}$ and $\mathcal{R}$ represent the sets of entities and relations, respectively. Each triple $\langle e, r, e' \rangle$ encodes a relationship $r$ between entity $e$ and entity $e'$. 

Given a natural language question $Q$ that contains a topic entity $e_{\text{topic}}$, the goal is to identify an answer entity $e_{\text{ans}} \in \mathcal{E}$ within $\mathcal{G}$ that correctly answers $Q$. Unlike simple KBQA, where $e_{\text{ans}}$ is directly connected to $e_{\text{topic}}$, complex KBQA requires multi-hop reasoning, as $e_{\text{ans}}$ may be multiple hops away from $e_{\text{topic}}$.

\section{Method}

Our method, shown in Figure~\ref{fig:main} and detailed in Algorithm~1, consists of three modules: (1) \textit{Iterative Question Guidance}, (2) \textit{Two-hop Entity Exploration}, and (3) \textit{Answer Extraction}. Given a complex query, \textit{Iterative Question Guidance} breaks it down into simpler sub-questions, each guiding the next exploration step. \textit{Two-hop Entity Exploration} expands relevant entities and relations by leveraging both 1-hop and 2-hop connections. Finally, \textit{Answer Extraction} integrates the retrieved knowledge to generate the final answer.

\begin{algorithm}[h]
\caption{Question-Guided KBQA}
\label{alg:kbqa}
\KwIn{
    Natural language question $Q$; 
    Topic entity $e_{\text{topic}}$; 
    Knowledge graph $\mathcal{G}$.
}
\KwOut{Answer $A$ to the question $Q$.}

\textbf{Initialization:} \\
$\mathcal{C} \leftarrow \emptyset$ \hfill // context \\
$\mathcal{E} \leftarrow \{\,e_{\text{topic}}\}$ \hfill // candidate entities \\

\For{$i \leftarrow 1$ \KwTo \text{MaxIter}}{
    \textbf{(1) Subquestion Generation:} \\
    $Q_{\text{sub}} \leftarrow \text{IGQ-LLM}(Q,\;\mathcal{C})$; \\

    \textbf{(2) Neighbor Retrieval:} \\
    $\mathrm{Nbr} \leftarrow \mathrm{SPARQL}(\mathcal{G},\;\mathcal{E})$; \\

    \textbf{(3) GNN-based Entity Selection:} \\
    \quad (a) Collect 2-hop neighbors: \\
    \quad $\mathcal{E}_{2h} \leftarrow \bigcup_{e \in \mathrm{Nbr}} \mathrm{SPARQL}(\mathcal{G},\;e)$; \\
    \quad (b) Relevance scoring: \\
    \quad $\mathrm{Scores} \leftarrow \mathrm{GNN}(\mathcal{E}_{2h},\;Q_{\text{sub}})$; \\
    \quad (c) Top-$k$ update: \\
    \quad $\mathcal{E} \leftarrow \text{Top-}k(\mathrm{Scores})$; \\

    \textbf{(4) Subquestion Answering:} \\
    $A_{\text{sub}} \leftarrow \text{AE-LLM}(Q_{\text{sub}},\;\mathcal{E})$; \\

    \textbf{(5) Context Update:} \\
    $\mathcal{C} \leftarrow \mathcal{C} \cup \{\,(Q_{\text{sub}},\,A_{\text{sub}})\}$; \\

    \textbf{(6) Sufficiency Check:} \\
    \If{ ${\operatorname{AE-LLM}} (\mathcal{C}) == \text{``sufficient''}$}{
        \textbf{break} \hfill // exit if enough evidence
    }
}

\Return $\text{AE-LLM}(Q,\;\mathcal{C})$ \hfill // final answer
\end{algorithm}

\subsection{Iterative Question Guidance}
Given a multi-step reasoning question $Q$, our goal is to dynamically decompose it into simpler sub-questions, guiding the knowledge graph exploration step by step. For instance, given the original question: \textit{"In the North Pacific region, what is the official flower of the affected area of the cyclone Tropical Storm Fabio?"}, our method first extracts the most relevant sub-question: \textit{"Which area in the North Pacific region was affected by Tropical Storm Fabio?"}.

To achieve this dynamic decomposition, we extract the next sub-question by considering both the reasoning context \(\mathcal{C}\) and the original question \(Q\).
Here, the context \(\mathcal{C}\) consists of previously generated sub-questions $Q_{\text{sub}}^{(i)}$ and their corresponding answers $A_{\text{sub}}^{(i)}$, providing a structured history of reasoning steps. This process is formally described in Step (1) of the Algorithm 1.
The extraction process can be formulated as:

\begin{equation}
Q_{\text{sub}}^{(n)} = \text{IQG-LLM} \Bigl(
    Q, \mathcal{C}
\Bigr) 
\end{equation}
where $\mathcal{C} = \bigl[
    Q_{\text{sub}}^{(1)}, A_{\text{sub}}^{(1)}, 
    Q_{\text{sub}}^{(2)}, \dots, 
    A_{\text{sub}}^{(n-1)}
\bigr]$.
\vspace{0.3em}

Following the Chain-of-Thought paradigm, we iteratively update prompts to reflect new sub-questions and answers. This structured, step-by-step prompting ensures coherent reasoning and preserves contextual consistency. At each step, the LLM also decides whether further decomposition is required. If not, the current sub-question is used to guide exploration of the knowledge graph, ensuring efficiency. 

\subsection{Two-Hop Based Entity Exploration} 
\textbf{Neighbor Retrieval.}~~
To retrieve relevant knowledge for each sub-question, we explore the neighbors of the target entity via SPARQL queries, as shown in Step~(2) of Algorithm~1. We employ a generic SPARQL template to extract 1-hop neighbors.
Since SPARQL query syntax of retrieving neighbors is fixed, we can reuse a template to systematically obtain them. For instance, if we wish to find all movies directed by Christopher Nolan, we can issue the following SPARQL query:


\begin{lstlisting}[style=sparql]
SELECT ?tailEntity
WHERE {
  ns:m.0bxtg ns:film.director.film ?tailEntity .
}
\end{lstlisting}
\noindent
where \texttt{ns:m.0bxtg} is the Freebase ID for Christopher Nolan, and \texttt{ns:film.director.film} is the relevant predicate.

We repeat this process to retrieve all outgoing edges and neighboring nodes for both head and tail entities in the current state of reasoning, ensuring a comprehensive collection of candidate neighbors.

Once we obtain the 1-hop neighbors of the current entity, we score each neighbor according to its relevance to the query. We then select the top $k$ neighbors as the next step in the reasoning process. 

\vspace{0.3em}
\noindent
\textbf{Two-Hop Neighbor Aggregation with GNN.} 
While 1-hop neighbors can be useful, they may not provide sufficient semantic context. 
To incorporate information from 2-hop neighbors, we perform another round of SPARQL queries to retrieve these nodes. We then employ a GNN to aggregate these 2-hop neighborhood signals, as shown in Step (3) in Algotithm 1. Specifically, we first convert textual features of each node (e.g., entity descriptions or relevant text) into dense vector representations using BERT. Let $\mathbf h_{1h}$ denote the representation for a 1-hop neighbor $e_{1h}$, while $\mathbf h_{2h}$ stands for a 2-hop neighbor $e_{2h}$. We adopt a GraphSAGE-like aggregation~\cite{hamilton2017inductive} to update the representation of each 1-hop neighbor with its 2-hop neighbors:
\begin{equation}
\label{eq:gnn_aggregation}
\hat{\mathbf h}_{1h} = \sigma \Big(\mathbf W \cdot [\mathbf h_{1h}|| \text{AGG}\{\mathbf h_{2h} \mid e_{2h} \in \mathcal{N}(e_{1h})\}]\Big),
\end{equation}
where $\mathcal{N}(e_{1h})$ denotes the set of neighbor nodes of $e_{1h}$, $\mathbf W$ is a trainable weight matrix, and $\text{AGG}(\cdot)$ is an aggregation function such as mean-pooling, which can aggregate neighboring nodes without any numerical limitations. By concatenating the central node's representation $\mathbf h_{1h}$ with the aggregated neighbor representation $\mathbf h_{2h}$, we preserve the original information of the central node and mitigate the risk of diluting its identity.

\vspace{0.3em}
\noindent
\textbf{Relevance Classification.}~~
After updating the representation of each 1-hop neighbor to $\hat{\mathbf h}_{1h}$, we concatenate it with the sub-question representation $Q_{\text{sub}}$, forming a combined vector $\mathbf{h}$. We then feed $\mathbf{h}$ through a two-layer MLP to perform a binary classification (relevant vs.\ irrelevant). Formally,
\begin{equation}
\label{eq:mlp_classification}
\hat{\mathbf{y}} = \mathrm{Softmax}\bigl(\mathbf{W}_2 \,\sigma(\mathbf{W}_1 \,\mathbf{h} + \mathbf{b}_1) + \mathbf{b}_2\bigr),
\end{equation}
where $\mathbf{W}_1, \mathbf{W}_2$ are weight matrices, $\mathbf{b}_1, \mathbf{b}_2$ are biases, and $\sigma(\cdot)$ is a nonlinear activation function such as ReLU. The Softmax function ensures that the output $\hat{\mathbf{y}}$ represents the probability of the neighbor being relevant or irrelevant.
To obtain a normalized relevance score, we take the probability corresponding to the "relevant" class:
\begin{equation}
    score = \hat{\mathbf{y}}[1] \in (0,1).
\end{equation}

Finally, we select the top-$k$ entities with the highest scores as the supporting evidence for answering the question.

We use cross-entropy loss with one-hot encoding for training:
\begin{equation}
L = -\sum_{i=1}^{2} y_i \log(\hat{y}_i).
\end{equation}
Here, $y_i$ is the ground-truth label for class $i$, and $\hat{y}_i$ is the predicted probability of class $i$. 


\subsection{Answer Extraction LLM (AE-LLM)}
After selecting the top-$k$ most relevant entities, we use an LLM to answer the current sub-question, following steps (4)–(6) in Algorithm 1. Specifically, we construct a prompt instructing the LLM to generate an answer based on both the retrieved knowledge and its internal knowledge. The generated answer is then incorporated into the context $\mathcal{C}$ along with the sub-question. Next, we assess whether $\mathcal{C}$ contains sufficient information to directly answer the original question. If so, we prompt the LLM to synthesize all sub-questions and answers to produce the final response. Otherwise, we iteratively generate additional sub-questions, retrieve relevant entities, and obtain answers until enough information is accumulated.

\section{Experiments}

\begin{table*}[h]
    \centering
    \resizebox{0.90\textwidth}{!}{ 
    \begin{tabular}{l|ccc|c}
        \hline
        \multirow{2}{*}{Method} & \multicolumn{3}{c|}{Multi-hop reasoning} & Generalization \\
        \cline{2-5}
        & CWQ & WebQSP & WebQuestion & GrailQA \\
        \hline
        Question-Decomposion~\cite{huang2023question}  & 72.8 & - & - & - \\
        Chain-of-Question~\cite{yixing2024chain}      & \textbf{78.8} & 78.10 & - & - \\
        KG-CoT~\cite{zhao2024kg}                      & 62.3 & \underline{84.90} & \underline{68.00} & - \\
        FlexKBQA~\cite{li2024flexkbqa}               & - & 46.20 & - & 68.90 \\
        Interactive-KBQA~\cite{xiong-etal-2024-interactive} & 49.07 & 71.2 & - & - \\
        ToG~\cite{sun2024thinkongraph}               & 69.5 & 82.6 & 57.90 & \textbf{81.4} \\
        \hline
        \textbf{iQUEST (GPT-4o based)}            & \underline{73.85} & \textbf{88.93} & \textbf{81.23} & \underline{73.52} \\
        \hline
    \end{tabular}
    } 
        \caption{Performance comparison of different methods. 
    \textbf{Best results} in bold and \underline{second-best} in underline. All the comparison results are taken from their corresponding papers.}
    \label{tab:comparison}
\end{table*}

\paragraph{Datasets}~
We evaluate our approach on four standard KBQA datasets, all of which require multi-hop reasoning: ComplexWebQuestions (CWQ)~\cite{talmor2018web}, WebQuestionsSP (WebQSP)~\cite{yih2016value}, WebQuestions~\cite{berant2013semantic}, and GrailQA~\cite{gu2021beyond}. Among them, GrailQA is also used to assess the generalization ability of the retrieval model. All these datasets are based on the Freebase knowledge graph~\cite{bollacker2008freebase}.
Following previous studies, we adopt Hit@1 score as the primary evaluation metric.

\textbf{Implementation Details.}
For the LLMs, we utilize GPT-4o (\texttt{gpt-4o-2024-05-13}), one of the most advanced general-purpose LLMs, alongside DeepSeek-R1 (70B), a recently released, powerful open-source reasoning model~\cite{guo2025deepseek}. Additionally, we incorporate LLaMA 3.1-70B (hereafter referred to as LLaMA 70B) and a smaller-scale model, LLaMA 3.2-3B (referred to as LLaMA 3B), for comparative analysis.  

For the GNN-based reasoning module, we use \texttt{bert-base-uncased} as the encoder, with a hidden dimension of 768. The GNN itself has a hidden dimension of 128. During inference, we retrieve the top-3 most relevant entities as supporting evidence to answer the question. Since the GNN evaluates the relevance between entities and the question independently at each inference step, we train its parameters using single-hop reasoning samples from the training datasets. Negative samples are generated through random negative sampling.

\subsection{Main results}
Table \ref{tab:comparison} presents our experimental results across four standard KBQA benchmarks. Our iQUEST achieves state-of-the-art performance on WebQSP and WebQuestions and ranks second on CWQ and GrailQA. Compared to baselines that do not fine-tune LLMs, such as ToG and Interactive-KBQA, our approach demonstrates superior reasoning over large knowledge graphs. Interactive-KBQA employs Mistral-13B, whose relatively small size limits performance, while ToG benefits from GPT-4 through an API. Despite these differences, our method significantly outperforms both, underscoring its effectiveness without requiring LLM fine-tuning.

For the GrailQA dataset (64k samples), ToG randomly selected 1,000 instances to reduce computation, and we followed the same strategy for fairness, though results may vary slightly due to different samples. On WebQuestions, where ambiguity is common in real user queries, our approach uses the original question as guidance, achieving a 23\% performance gain over ToG.

Among fine-tuned approaches, FlexKBQA, KG-CoT, Chain-of-Question, and Question-Decomposition deliver competitive results. Chain-of-Question leverages GPT-3.5-turbo for reasoning and a fine-tuned T5 model for question decomposition, achieving strong performance. Nevertheless, our method maintains an advantage across most datasets. Overall, iQUEST excels in multi-hop reasoning and generalization. Instead of fine-tuning an LLM, it trains a GNN for entity exploration, enhancing adaptability while simplifying training.

\subsection{Ablation Study}

\begin{table*}[ht!]
    \centering
    \renewcommand{\arraystretch}{1.0}
    \resizebox{0.80\textwidth}{!}{
    \begin{tabular}{lcccc}
        \toprule
        \multirow{2}{*}{Model Combination} & \multicolumn{4}{c}{Dataset} \\
        \cmidrule(lr){2-5}
        & CWQ & WebQSP & WebQuestion & GrailQA \\
        \midrule
        LLaMA 3B (AE) + GPT-4o (IQG) & 20.14 & 40.42 & 48.65 & 32.87 \\
        LLaMA 3B (AE) + GPT-4o (IQG) + GNN & 23.66 \textbf{(+3.52)} & 43.73 \textbf{(+3.31)} & 50.11 \textbf{(+1.46)} & 34.19 \textbf{(+1.32)} \\
        \midrule
        LLaMA 70B (AE) + GPT-4o (IQG) & 47.24 & 83.07 & 74.61 & 58.46 \\
        LLaMA 70B (AE) + GPT-4o (IQG) + GNN & 50.30 \textbf{(+3.06)} & 84.36 \textbf{(+1.29)} & 76.65 \textbf{(+2.04)} & 60.45 \textbf{(+1.99)} \\
        \midrule
        DeepSeek-R1 (AE) + GPT-4o (IQG) & 52.44 & 80.43 & 74.63 & 58.23 \\
        DeepSeek-R1 (AE) + GPT-4o (IQG) + GNN & 55.64 \textbf{(+3.20)} & 83.21 \textbf{(+2.78)} & 77.87 \textbf{(+3.24)} & 60.27 \textbf{(+2.04)} \\
        \midrule
        GPT-4o (AE) + GPT-4o (IQG) & 68.42 & 88.10 & 80.20 & 69.30 \\
        GPT-4o (AE) + GPT-4o (IQG) + GNN & 73.85 \textbf{(+5.43)} & 88.93 \textbf{(+0.83)} & 81.23 \textbf{(+1.03)} & 73.52 \textbf{(+4.22)} \\
        \bottomrule
    \end{tabular}}
    \vspace{-0.2cm}
    \caption{GNN-based search across AE-LLMs and datasets.}
    \label{tab:e1_2}
\end{table*}

\begin{table*}[ht!]
    \centering
    \renewcommand{\arraystretch}{1.0}
    \resizebox{0.80\textwidth}{!}{
    \begin{tabular}{lcccc}
        \toprule
        \multirow{2}{*}{Model Combination} & \multicolumn{4}{c}{Dataset} \\
        \cmidrule(lr){2-5}
        & CWQ & WebQSP & WebQuestion & GrailQA \\
        \midrule
        GPT-4o (No IQG - Baseline) & 63.34 & 84.25 & 77.72 & 67.42 \\
        GPT-4o (AE) + LLaMA 3B (IQG) & 64.95 {(+1.61)} & 85.70 {(+1.45)} & 78.46 {(+0.74)} & 67.69 {(+0.27)} \\
        GPT-4o (AE) + LLaMA 70B (IQG) & 67.46 {(+4.12)} & 87.01 {(+2.76)} & 79.78 {(+2.06)} & 68.76 {(+1.34)} \\
        GPT-4o (AE) + DeepSeek-R1 (IQG) & 68.16 {(+4.82)} & 87.85 {(+3.60)} & 80.11 {(+2.39)} & 69.45 {(+2.03)} \\
        GPT-4o (AE) + GPT-4o (IQG) & 68.42 {(+5.08)} & 88.10 {(+3.85)} & 80.20 {(+2.48)} & 69.30 {(+1.88)} \\
        GPT-4o (AE) + GPT-4o (IQG) + GNN & 73.85 {(+10.51)} & 88.93 {(+4.68)} & 81.23 {(+3.51)} & 73.52 {(+6.10)} \\
        \bottomrule
    \end{tabular}
    }
    \vspace{-0.2cm}
    \caption{Performance differences compared to GPT-4o (No IQG as baseline) across datasets.}
    \label{tab:e2_1}
    \vspace{-0.4cm}
\end{table*}

\subsubsection{Assumptions}
\label{sec:assumptions}
Ablation studies on LLM systems are challenging due to their black-box nature. 
To analyze the impact of our design choices, we explicitly outline our assumptions based on prior knowledge and empirical studies presented in the literature:

\noindent \textbf{A1.} The most relevant capabilities of an LLM in terms of QA tasks are  \emph{reasoning} and \emph{internal knowledge}.

\noindent \textbf{A2.} LLMs with significantly smaller parameter sizes generally exhibit lower overall capability. For example, LLaMA 3B performs worse than LLaMA 70B and DeepSeek-R1 (70B), while GPT-4o is considered the most capable LLM.

\noindent \textbf{A3.} We assume that model size serves as a sufficient proxy for internal knowledge. That is, LLMs of similar size tend to have comparable levels of internal knowledge.

\noindent \textbf{A4.} Based on prior empirical evidence and available information, we assume that DeepSeek-R1 surpasses LLaMA-70B in reasoning due to its reasoning-focused pretraining, and DeepSeek-R1 has at least comparable reasoning capability to GPT-4o.
we summarize our assumptions about each LLM in the following Table.
\begin{center}
    \renewcommand{\arraystretch}{1.0}
    \resizebox{0.48\textwidth}{!}{
    \begin{tabular}{lccc}
      \hline
      \toprule
      Model & Reasoning & Internal Knowledge & Overall\\
            \midrule
        LLaMA 3B & Low & Low & Low\\
        LLaMA 70B & Medium & Medium& Medium \\
        DeepSeek-R1 & High & Medium & Medium\\
        GPT-4o & High & High & High\\
        \bottomrule
    \end{tabular}
    }
\end{center}
\subsubsection{Experimental Design}
In this paper, we introduce three key components (C1-C3) and conduct corresponding ablations to analyze their impact. The complete ablation results can be found in Table~\ref{tab:qa_combination} (Appendix~\ref{sec:appendix}).

\paragraph{C1. KG Augmentation}
We integrate a KG with a GNN-based forward-looking retrieval mechanism to complement the LLM's internal knowledge. We investigate the following aspects:

\begin{itemize}
    \vspace{-2pt}
    \item \textbf{C1.1 KG effectiveness:} Prior studies show that KGs effectively enrich knowledge. While this is not our primary focus, we conduct an ablation where we exclude the KG to replicate this conclusion.
    \vspace{-3pt}
    \item \textbf{C1.2 GNN-based retrieval:} Retrieving relevant neighbors from the KG is challenging due to the abundance of semi-relevant nodes. Excluding too many may omit crucial information, while including too many introduces noise and reduces efficiency. We evaluate the effectiveness of our GNN-based forward-looking two-hop neighbor retrieval model in balancing this trade-off.
\end{itemize}

\vspace{-4pt}
\paragraph{C2. Iterative Question Guidance (IQG-LLM)}
IQG-LLM serves as an interface between the KG and the answer-generation LLM, effectively offloading the reasoning burden. To evaluate its impact, we conduct the following ablations:

\begin{itemize}
    \vspace{-2pt}
  \item \textbf{C2.1 Removing IQG-LLM:} We assess the impact of directly generating answers without this intermediate reasoning module.
  \vspace{-2pt}
  \item \textbf{C2.2 Reasoning importance:} We hypothesize that IQG-LLM primarily enhances reasoning. To test this, we examine whether stronger reasoning LLMs perform better in this role. Specifically, if the IQG-LLM has strong reasoning capabilities while the AE-LLM does not, the IQG-LLM should improve the AE-LLM’s QA performance more than an IQG-LLM with weaker reasoning abilities.
\end{itemize}

\vspace{-5pt}
\paragraph{C3. Answer Extraction LLM (AE-LLM)}
AE-LLM takes the question and retrieved context as input and is responsible for understanding and summarizing relevant information. We explore the following hypotheses:

\begin{itemize}
    \item \textbf{C3.1 Internal knowledge vs reasoning requirement:} We evaluate the extent to which AE-LLM relies on its internal knowledge vs reasoning. Our hypothesis is that if IQG-LLM provides strong reasoning supports, the need for AE-LLM to independently reason is significantly reduced. However, the internal knowledge still plays an important role.
\end{itemize}

\subsubsection{Results}
In this section, we examine the impact of these components by systematically removing or altering them. Given the assumptions presented in Section~\ref{sec:assumptions}, we present our ablation results for each component C1–C3 in detail.

\begin{table*}[ht!]
    \centering
    \renewcommand{\arraystretch}{1.0}
    \resizebox{0.70\textwidth}{!}{
    \begin{tabular}{lcccc}
        \toprule
        \multirow{2}{*}{IQG-LLM} & \multicolumn{4}{c}{Dataset} \\
        \cmidrule(lr){2-5}
        & CWQ & WebQSP & WebQuestion & GrailQA \\
        \midrule
        DeepSeek-R1 (baseline) & 46.15 & 82.64 & 73.41 & 57.61 \\
        LLaMA 70B (weaker reasoning) & 47.13 (+0.98) & 83.15 (+0.51) & 74.36 (+0.95) & 58.57 (+0.96) \\
        GPT-4o (equivalent reasoning) & 47.24 (-0.11) & 83.07 (+0.08) & 74.61 (-0.25) & 58.46 (+0.11) \\
        \bottomrule
    \end{tabular}
  }
  \vspace{-0.2cm}
    \caption{Performance comparison for LLaMA 70B as AE-LLM across different IQG-LLM models. Numbers in the parentheses indicate relative change from DeepSeek-R1 as AE-LLM (baseline).}
    \label{tab:e2_2}
    \vspace{-0.1cm}
\end{table*}

\begin{table*}[t!]
    \centering
    \renewcommand{\arraystretch}{1.0}
    \resizebox{0.74\textwidth}{!}{
    \begin{tabular}{lcccc}
        \toprule
        \multirow{2}{*}{AE-LLM} & \multicolumn{4}{c}{Dataset} \\
        \cmidrule(lr){2-5}
         & CWQ & WebQSP & WebQuestion & GrailQA \\
        \midrule
        DeepSeek-R1 (baseline) & 55.64 & 83.21 & 77.87 & 60.27 \\
        LLaMA 70B (weaker reasoning) & 50.3 (-5.34) & 84.36 (+1.15) & 76.65 (-1.22) & 60.45 (+0.18) \\
      GPT-4o (stronger internal knowledge) & 73.85 (+18.21) & 88.93 (+5.72) & 81.23 (+3.36) & 73.52 (+13.25) \\
        \bottomrule
    \end{tabular}
  }
  \vspace{-0.2cm}
    \caption{Impact of AE-LLM selection on QA performance using GPT-4o as IQG-LLM with GNN-based retrieval. Numbers in the parentheses indicate relative change from DeepSeek-R1 as AE-LLM (baseline).}
    \label{tab:e3}
    \vspace{-0.4cm}
\end{table*}

\subsubsection*{C1.1 KG Effectiveness}
To validate the importance of KG augmentation, we remove the KG and observe its impact across different models (rows {without KG} vs {with KG}). The results in Table~\ref{tab:qa_combination} show a substantial drop in performance. For example, GPT-4o (without KG) achieves only 40.14\% on CWQ, compared to 63.34\% with KG, indicating that even the strongest LLMs significantly benefit from structured external knowledge. This can be observed across all datasets for larger LLMs (i.e., all LLMs except for LLaMA 3B). Interestingly, for LLaMA 3B, the QA capability decreases when the KG is present. This suggests that smaller models may struggle to effectively integrate external knowledge and may even be hindered by conflicting or irrelevant KG information.


\subsubsection*{C1.2 GNN-based Two-Hop Neighbor Retrieval}
Our GNN-based retrieval model addresses the trade-off between retrieving too many semi-relevant nodes (introducing noise) and excluding too many (losing critical information by looking one step ahead and incorporating semantic information from second-hop neighbors.
Table~\ref{tab:e1_2} demonstrates that this mechanism consistently improves performance across all datasets and model configurations.
Even GPT-4o (AE) + GPT-4o (IQG), the strongest model pairing, benefits from GNN-based search, particularly on CWQ (+5.43\%) and GrailQA (+4.22\%). This highlights the importance of fine-grained retrieval even for highly capable models, particularly for complex multi-hop reasoning and generalization tasks. While stronger AE-LLMs already perform well, they still leverage improved knowledge selection to refine answer accuracy.

\subsubsection*{C2.1 Removing IQG-LLM}
IQG-LLM plays a crucial role in structuring multi-hop reasoning by breaking complex questions into intermediate steps. As shown in Table~\ref{tab:qa_combination}, removing IQG-LLM leads to performance drops across all datasets. Particularly, we illustrate this for the most capable LLM GPT-4o in Table~\ref{tab:e2_1}. Even using a much smaller LLaMA 3B as IQG-LLM increases the QA performance.

\subsubsection*{C2.2 Reasoning Importance in IQG-LLM}

We expect AE-LLMs with weaker reasoning abilities, such as LLaMA 70B, to gain more from a reasoning-capable IQG-LLM. Given our assumption that DeepSeek-R1 has stronger reasoning than LLaMA 70B and is at least as capable as GPT-4o in reasoning, while GPT-4o has superior internal knowledge and general capabilities, we analyze how different IQG-LLMs with varying reasoning strengths affect LLaMA 70B used as AE-LLM.
From Table~\ref{tab:e2_2} we see that using LLaMA 70B as AE-LLM, replacing DeepSeek-R1 with LLaMA 70B as IQG-LLM improves performance across all datasets. However, further upgrading to GPT-4o as IQG-LLM results in minimal changes, with slight improvements on WebQSP (\textbf{+0.08\%}) but minor drops on CWQ (\textbf{-0.11\%}) and WebQuestion (\textbf{-0.25\%}). These results indicate diminishing returns beyond a certain IQG-LLM reasoning threshold.

\subsubsection*{C3.1 Internal Knowledge Requirement in AE-LLM}
Table~\ref{tab:e3} examines the impact of AE-LLM selection on QA performance when using GPT-4o as IQG-LLM with GNN-based retrieval for the best reasoning and external knowledge support. Replacing DeepSeek-R1 with LLaMA 70B as AE-LLM results in only minor performance variations, suggesting that AE-LLM with weaker reasoning does not significantly affect the final QA outcome.
However, upgrading AE-LLM to GPT-4o leads to substantial improvements across all datasets. This indicates that AE-LLM benefits more from stronger internal knowledge rather than reasoning.

\section{Conclusion}
In this paper, we present iQUEST, an iterative, question-driven framework for multi-step reasoning in complex KBQA. It decomposes complex queries into sub-questions to guide structured reasoning, using a GNN for knowledge aggregation from relevant areas, which supports 2-hop forward-looking reasoning. By combining sub-questions and answers, iQUEST maintains a coherent reasoning path for more accurate results. Carefully designed ablation studies on four benchmark datasets and four LLMs demonstrate that generating question effectively reduces the reasoning burden, improving reasoning efficiency, knowledge extraction, and utilization. These results confirm the effectiveness of iQUEST in handling complex KBQA tasks.

\section*{Acknowledgment}
This work was partially funded by the Wallenberg AI, Autonomous Systems and Software Program (WASP), supported by Verket för innovationssystem (Vinnova), and the Chalmers Artificial Intelligence Research Centre (CHAIR). The computations in this work were enabled by resources provided by the National Academic Infrastructure for Supercomputing in Sweden (NAISS), partially funded by the Swedish Research Council through grant agreement no. 2022-06725.

\section*{Limitations}
Despite the effectiveness of our approach, several limitations should be acknowledged. First, our framework employs one LLM for question formulation and guidance, followed by another LLM for answering, which, while improving performance, introduces additional computational overhead. This increased cost may impact scalability, particularly in real-time or resource-constrained applications. We provide a detailed runtime comparison in Appendix~\ref{sec:runtime}

Second, our GNN model is limited to capturing information from 2-hop neighbors. While this is sufficient for many cases, it may be inadequate for certain domain-specific KGs where meaningful reasoning requires a broader context. Future work could explore more advanced path search strategies that allow consideration of multi-hop neighbors dynamically, potentially enhancing the model’s ability to leverage deeper relational structures within KGs.
\bibliography{custom}

\appendix

\section{Complete ablation results}
\label{sec:appendix}
Table \ref{tab:qa_combination} presents a comprehensive performance comparison of various model combinations in multi-hop reasoning and generalization tasks across four benchmark datasets: CWQ, WebQSP, WebQuestion, and GrailQA. The results highlight the effectiveness of different architectures and model integration strategies, particularly focusing on the impact of Knowledge Graphs (KG), Iterative Question Generation (IQG), Answer Extraction (AE) and Graph Neural Networks (GNN)-based information retrieval.  Additionally, the evaluation encompasses four large language models (LLaMA 3B, LLaMA 70B, DeepSeek-R1 (70B), and GPT-4o), comparing their ability to enhance reasoning capabilities and generalization performance.
\begin{table*}
    \centering
    \renewcommand{\arraystretch}{1}
    \resizebox{0.70\textwidth}{!}{
    \begin{tabular}{l|c|c|c|c}
        \toprule
        \textbf{Model Combination} & \multicolumn{3}{c|}{\textbf{Multi-hop Reasoning}} & \textbf{Generalization} \\
        & CWQ & WebQSP & WebQuestion & GrailQA \\
        \hline
        LLaMA 3B (without KG)  & 10.18 & 46.50 & 47.86 & 18.51 \\
        LLaMA 3B (with KG) & 14.81 & 30.51 & 40.00 & 29.06 \\
        LLaMA 3B (AE) + LLaMA 3B (IQG) & 16.65 & 31.97 & 41.56 & 29.73 \\
        LLaMA 3B (AE) + LLaMA 70B (IQG) & 19.44 & 37.74 & 46.58 & 31.29 \\
        LLaMA 3B (AE) + DeepSeek-R1 (IQG) & 19.86 & 38.56 & 48.11 & 32.75 \\
        LLaMA 3B (AE) + GPT-4o (IQG) & 20.14 & 40.42 & 48.65 & 32.87 \\
        \rowcolor{gray!30} LLaMA 3B (A) + GPT-4o + GNN & 23.66 & 43.73 & 50.11 & 34.19 \\
        \hline
        LLaMA 70B (without KG)  & 35.85 & 69.66 & 58.42 & 32.48 \\
        LLaMA 70B (with KG) & 42.65 & 80.43 & 71.11 & 56.35 \\
        LLaMA 70B (AE) + LLaMA 3B (IQG) & 43.33 & 81.85 & 72.15 & 56.43 \\
        LLaMA 70B (AE) + LLaMA 70B (IQG) & 46.15 & 82.64 & 73.41 & 57.61 \\
        LLaMA 70B (AE) + DeepSeek-R1 (IQG) & 47.13 & 83.15 & 74.36 & 58.57 \\
        LLaMA 70B (AE) + GPT-4o (IQG) & 47.24 & 83.07 & 74.61 & 58.46 \\
        \rowcolor{gray!30} LLaMA 70B (A) + GPT-4o + GNN & 50.30 & 84.36 & 76.65 & 60.45 \\
        \hline
        DeepSeek-R1 (without KG)  & 30.15 & 75.71 & 67.24 & 30.15 \\
        DeepSeek-R1 (with KG) & 50.85 & 78.79 & 74.01 & 57.68 \\
        DeepSeek-R1 (AE) + LLaMA 3B (IQG) & 51.02 & 79.13 & 74.18 & 57.91 \\
        DeepSeek-R1 (AE) + LLaMA 70B (IQG) & 51.78 & 79.43 & 74.35 & 58.14 \\
        DeepSeek-R1 (AE) + DeepSeek-R1 (IQG) & 52.05 & 79.08 & 74.56 & 58.05 \\
        DeepSeek-R1 (AE) + GPT-4o (IQG) & 52.44 & 80.43 & 74.63 & 58.23 \\
        \rowcolor{gray!30} DeepSeek-R1 + GPT-4o + GNN & 55.64 & 83.21 & 77.87 & 60.27 \\
        \hline
        GPT-4o (without KG) & 40.14 & 71.04 & 63.30 & 33.62 \\
        GPT-4o (with KG) & 63.34 & 84.25 & 77.72 & 67.42 \\
        GPT-4o (AE) + LLaMA 3B (IQG) & 64.95 & 85.70 & 78.46 & 67.69 \\
        GPT-4o (AE) + LLaMA 70B (IQG) & 67.46 & 87.01 & 79.78 & 68.76 \\
        GPT-4o (AE) + DeepSeek-R1 (IQG) & 68.16 & 87.85 & 80.11 & 69.45 \\
        GPT-4o (AE) + GPT-4o (IQG) & 68.42 & 88.10 & 80.20 & 69.30 \\
        \rowcolor{gray!30} GPT-4o (A) + GPT-4o(IQG) + GNN & 73.85 & 88.93 & 81.23 & 73.52 \\
        \bottomrule
    \end{tabular}
    }
    \caption{Performance comparison of different model combinations in multi-hop reasoning and generalization tasks.}
    \label{tab:qa_combination}
\end{table*}

\section{Comparison of Relative Runtime}
\label{sec:runtime}
\subsection{Overview}

We present a detailed runtime comparison between our \textit{iterative question-guided framework} and ToG \footnote{https://github.com/GasolSun36/ToG}, a representative baseline that also leverages LLMs for question decomposition and reasoning over knowledge graphs. Unlike our method, ToG does not incorporate explicit question-guided prompting, making it a suitable point of comparison in both efficiency and effectiveness.

\subsection{Evaluation Setup}

Both methods were evaluated under identical experimental settings using \texttt{GPT-4o} via Azure OpenAI. For each hop category (1-hop, 2-hop, and 3-hop), we randomly sampled 100 questions. The $n$-hop category indicates that the shortest path between the topic and answer entities in the knowledge graph is $n$.

\subsection{Overall Performance Comparison}

Table~\ref{tab:runtime_full} summarizes the results in terms of the average number of LLM calls, runtime per query (in seconds), and accuracy.

\begin{table*}[h]
\centering
\renewcommand{\arraystretch}{1.2}
\begin{tabular}{c|c|c|c|c}
\toprule
\textbf{Hop} & \textbf{Method} & \textbf{LLM Calls} & \textbf{Runtime (s)} & \textbf{Hit@1} \\
\midrule
\multirow{2}{*}{1-hop} & ToG   & 2.32 & 3.41 & 87.0 \\
      & iQUEST  & 3.09 (+0.77) & 4.65 (+1.24) & 88.0 (+1.0) \\
\hline
\multirow{2}{*}{2-hop} & ToG   & 2.56 & 3.96 & 72.0 \\
      & iQUEST  & 5.86 (+3.30) & 7.54 (+3.58) & 78.0 (+6.0) \\
\hline
\multirow{2}{*}{3-hop} & ToG   & 3.88 & 6.03 & 37.0 \\
      & iQUEST  & 7.93 (+4.05) & 10.69 (+4.66) & 46.0 (+9.0) \\
\bottomrule
\end{tabular}
\caption{Overall runtime and accuracy comparison between ToG and our method.}
\label{tab:runtime_full}
\end{table*}

Our approach incurs higher latency due to increased LLM calls, particularly for complex multi-hop queries. However, the added reasoning steps improve the completeness of path exploration, yielding significantly better accuracy, especially in 2-hop and 3-hop settings. In contrast, ToG frequently terminates prematurely in harder cases, sacrificing accuracy for speed.

\subsection{Controlled Runtime Analysis (Correct Chains Only)}

To isolate runtime differences independent of correctness variance, we conducted an additional analysis limited to cases where both methods successfully produced complete and correct reasoning chains. Results are shown in Table~\ref{tab:runtime_correct}.

\begin{table*}[h]
\centering
\renewcommand{\arraystretch}{1.2}
\begin{tabular}{c|c|c|c}
\toprule
\textbf{Hop} & \textbf{Method} & \textbf{LLM Calls} & \textbf{Runtime (s)} \\
\midrule
\multirow{2}{*}{1-hop} & ToG    & 2.51             & 3.60 \\
                       & iQUEST & 3.12 (+0.61)     & 4.82 (+1.22) \\
\hline
\multirow{2}{*}{2-hop} & ToG    & 3.65             & 5.23 \\
                       & iQUEST & 6.05 (+2.40)     & 8.06 (+2.83) \\
\hline
\multirow{2}{*}{3-hop} & ToG    & 4.87             & 7.36 \\
                       & iQUEST & 8.32 (+3.45)     & 11.04 (+3.68) \\
\bottomrule
\end{tabular}
\caption{Runtime comparison on correctly answered cases only.}
\label{tab:runtime_correct}
\end{table*}

The difference in LLM calls aligns with our method’s structure: it introduces an explicit question-guided step at each reasoning hop, resulting in approximately 3 (1-hop: 1 questioning + 1 reasoning + 1 summary), 5 (2-hop), and 7 (3-hop) calls. In contrast, ToG performs around 2, 3, and 4 calls respectively. Empirical averages match the theoretical design, with our method using 0.81, 2.20, and 3.15 more calls per query for 1-hop, 2-hop, and 3-hop questions, respectively.

\section{Handling Missing or Incomplete Knowledge Graph Facts}

In real-world scenarios, knowledge graphs (KGs) are often incomplete or contain outdated information. To address this challenge, our system employs two complementary strategies to ensure robust question answering despite such limitations:

\begin{enumerate}
    \item \textbf{Leveraging LLM Internal Knowledge:} \\
    When the required fact is not available in the KG, \textsc{iQUEST} defers to the pretrained knowledge of the large language model (LLM) by directly querying it. This fallback mechanism proves effective in approximately half of such cases, as demonstrated in the example provided in Response~2.
    
    \item \textbf{Retrieving Indirect Supporting Facts:} \\
    In cases where the exact fact is missing, \textsc{iQUEST} searches for semantically related facts that can provide indirect support. For example, even if the exact timezone of a location is not found, knowing that \textit{Utah} is in the \textit{United States} allows the model to infer a plausible timezone.
\end{enumerate}
Despite employing these strategies, their effectiveness remains limited. In the following section, we provide a detailed analysis of the impact of missing or incorrect knowledge graph facts.

\section{Error Analysis: Missing Entities in the Knowledge Graph}

Some sub-questions fail during the reasoning process due to missing entities or incomplete evidence in the knowledge graph (KG). For instance, for the question \emph{"About the school that publishes the Harvard Review, what are its school colors?"}, the generated sub-question retrieves only an unnamed entity ID (\texttt{m.01066g18}), which lacks sufficient semantic information to support further reasoning steps.

To better understand the impact of such failures, we manually analyzed and categorized errors based on whether the relevant entity is missing in the KG or exists but lacks a semantic name. The results, broken down by dataset and whether iQUEST answered the question correctly or incorrectly, are summarized in Table~\ref{tab:missing_entity_analysis}.

\begin{table*}[h]
\centering
\caption{Percentage of sub-question failures due to missing or unnamed entities in the KG.}
\label{tab:missing_entity_analysis}
\renewcommand{\arraystretch}{1.1}
\setlength{\tabcolsep}{6pt}
\begin{tabular}{l|cc|cc}
\toprule
\textbf{Dataset} & \multicolumn{2}{c|}{\textbf{Entity Missing in KG}} & \multicolumn{2}{c}{\textbf{Entity Exists but Name Missing}} \\
 & \textbf{Correct} & \textbf{Incorrect} & \textbf{Correct} & \textbf{Incorrect} \\
\midrule
CWQ          & 12\% & 17\% & 6\%  & 13\% \\
WebQSP       & 6\%  & 4\%  & 2\%  & 4\%  \\
WebQuestion  & 3\%  & 5\%  & 1\%  & 6\%  \\
GrailQA      & 3\%  & 5\%  & 14\% & 9\%  \\
\bottomrule
\end{tabular}
\end{table*}

These results show that while LLMs like iQUEST are sometimes able to answer questions using prior knowledge, the incompleteness of the KG, either through missing entities or insufficient semantic labeling, remains a significant obstacle. Addressing this issue is critical for improving the robustness and generalization of KGQA systems.

\section{Failure Case Analysis}

\textbf{Original Question:} \\
\emph{Which child actor played in the movie whose soundtrack is Forrest Gump: Original Motion Picture?}

\vspace{0.5em}
\textbf{Reasoning Process via Sub-question Decomposition:} \\
To answer this complex question using a knowledge graph, the system first attempts to decompose it into simpler sub-questions. The goal is to resolve intermediate entities (such as the movie) before identifying the actor. The process unfolds as follows:

\begin{enumerate}
    \item \textbf{Sub-question 1:} \emph{What movie is associated with the soundtrack Forrest Gump: Original Motion Picture Score?} \\
    $\rightarrow$ No direct answer found in the knowledge graph.

    \item \textbf{Sub-question 2:} \emph{What is the soundtrack title of the movie Forrest Gump?} \\
    $\rightarrow$ No direct answer found.

    \item \textbf{Sub-question 3:} \emph{What movie has the soundtrack titled Forrest Gump: Original Motion Picture?} \\
    $\rightarrow$ No direct answer found.

    \item \textbf{Sub-question 4:} \emph{What movie features the soundtrack titled Forrest Gump: Original Motion Picture?} \\
    $\rightarrow$ Still no answer retrieved.

    \item \textbf{Sub-question 5:} \emph{What movies did Asa Earl Carter act in?} \\
    $\rightarrow$ This sub-question appears off-topic and indicates potential drift in the reasoning path.

    \item \textbf{Sub-question 6:} \emph{What is the movie associated with the soundtrack Forrest Gump: Original Motion Picture?} \\
    $\rightarrow$ No answer obtained, despite being a paraphrase of earlier attempts.

    \item \textbf{Sub-question 7:} \emph{What movie has the soundtrack titled ``Forrest Gump: Original Motion Picture''?} \\
    $\rightarrow$ No result returned.
\end{enumerate}

\vspace{0.5em}
\textbf{Final Answer Synthesis:} \\
Despite no explicit answer found in the knowledge graph for any sub-question, the system leverages world knowledge to synthesize an answer. It recognizes that \emph{Forrest Gump: Original Motion Picture} is the official soundtrack of the movie \emph{Forrest Gump}.

Further, among the limited child actors featured in \emph{Forrest Gump}, \textbf{Haley Joel Osment} is identified as the actor who portrayed Forrest Gump’s son in a brief but notable appearance. While not a main character, he is the only prominent child actor in the film.

\vspace{0.5em}
\textbf{Answer:} \\
\emph{Haley Joel Osment}

\vspace{0.5em}
\textbf{Analysis:} \\
This failure case highlights a key limitation of our method: when a generated sub-question fails to retrieve an answer from the knowledge graph, the reasoning process may stall or fall into repetitive, unproductive loops. This leads to the generation of redundant sub-questions and ineffective reasoning paths.

Moreover, retrieving irrelevant information from the graph can mislead the LLM to generate off-topic sub-questions, such as "\textit{What movies did Asa Earl Carter act in?}", which are unrelated to the original question.

These observations underscore a major challenge: the incompleteness of the knowledge graph remains a significant bottleneck for our method’s effectiveness.

\end{document}